\documentclass[runningheads]{llncs}
\usepackage[T1]{fontenc}
\usepackage{graphicx,verbatim}
\usepackage{amsmath, amssymb}
\usepackage{booktabs}
\usepackage{hyperref}
\usepackage{color}

\usepackage[acronym]{glossaries}
\newacronym{acdc}{ACDC}{Automated Cardiac Diagnosis Challenge}
\newacronym{mm2}{M\&Ms-2}{ Multi-Centre, Multi-View, Multi-Vendor \& Multi-Disease Cardiac Image Segmentation Challenge}

\begin{document}
\title{CardioDiT: Latent Diffusion Transformers for \\  4D Cardiac MRI Synthesis }
\titlerunning{CardioDiT: 4D Cardiac MRI Synthesis}
%
  \author{
  Marvin Seyfarth\inst{1,2,3} \and
  Sarah Kaye Müller\inst{1,2,3} \and
  Arman Ghanaat\inst{1} \and
  Isabelle Ayx\inst{4} \and
  Fabian Fastenrath\inst{5} \and
  Philipp Wild\inst{6} \and
  Alexander Hertel\inst{4} \and
  Theano Papavassiliu\inst{3,7}\and
  Salman Ul Hassan Dar\inst{1,2,3}\and
  Sandy Engelhardt\inst{1,2,3}
  }

  \institute{
  Institute for Artificial Intelligence in Cardiovascular Medicine, Medical Faculty of
Heidelberg University, Heidelberg University, Germany \and
  Department of Cardiology, Angiology, Pneumology, Heidelberg University Hospital,
Heidelberg, Germany \and
  DZHK (German Centre for Cardiovascular Research), Partner Site
Heidelberg/Mannheim, Heidelberg, Germany \and
  Department of Radiology and Nuclear Medicine, University Medical Center
Mannheim, Germany \and
Section for Invasive Cardiology and Electrophysiology, I. Medical Department,
Cardiology, Hemostasiology and Intensive Care, University Medical Centre
Mannheim, Mannheim, Germany \and
University Medical Center of the Johannes Gutenberg-University Mainz, Germany \and
Department of Cardiology, Angiology, Hemostasis, and Medical Intensive Care,
University Medical Centre Mannheim, Medical Faculty Mannheim, University of
Heidelberg, Germany\\
  \email{Marvin.Seyfarth@med.uni-heidelberg.de}
  }
 
 \authorrunning{M. Seyfarth et al.}


  
\maketitle              
\begin{abstract}
Latent diffusion models (LDMs) have recently achieved strong performance in 3D medical image synthesis. However, modalities like cine cardiac MRI (CMR), representing a temporally synchronized 3D volume across the cardiac cycle, add an additional dimension that most generative approaches do not model directly. Instead, they factorize space and time or enforce temporal consistency through auxiliary mechanisms such as anatomical masks. Such strategies introduce structural biases that may limit global context integration and lead to subtle spatiotemporal discontinuities or physiologically inconsistent cardiac dynamics. We investigate whether a unified 4D generative model can learn continuous cardiac dynamics without architectural factorization. We propose CardioDiT, a fully 4D latent diffusion framework for short-axis cine CMR synthesis based on diffusion transformers. A spatiotemporal VQ-VAE encodes 2D+t slices into compact latents, which a diffusion transformer then models jointly as complete 3D+t volumes, coupling space and time throughout the generative process. We evaluate CardioDiT on public CMR datasets and a larger private cohort, comparing it to baselines with progressively stronger spatiotemporal coupling. Results show improved inter-slice consistency, temporally coherent motion, and realistic cardiac function distributions, suggesting that explicit 4D modeling with a diffusion transformer provides a principled foundation for spatiotemporal cardiac image synthesis. Code and models trained on public data are available at \url{https://github.com/Cardio-AI/cardiodit}.
\end{abstract}

\section{Introduction}
Generative models, particularly latent diffusion models (LDMs), have recently shown strong performance in synthesizing high-quality 3D medical images \cite{Khader2023,wavelet_dm,meddiffusion}, enabling applications such as data augmentation or privacy-preserving data sharing where annotated datasets are scarce. Some clinically relevant modalities, like short-axis cardiac MRI (CMR), include a temporal component, meaning both spatial anatomy and temporal dynamics are critical for capturing cardiac function. Most existing generative approaches do not model the full 3D+t distribution directly. Instead, they factorize space and time, e.g., generating 2D slices with temporal modules \cite{texdc}, synthesizing 3D volumes at selected timepoints \cite{timepoint_syn}, or enforcing temporal consistency via auxiliary components \cite{maskedcmr}. While such strategies can produce visually plausible results, they treat space and time as partially separable components. This structural bias may limit global context integration and can result in subtle spatiotemporal inconsistencies, such as slice-wise discontinuities or anatomically implausible cardiac dynamics across the cycle. Consequently, modeling truly continuous 4D cardiac dynamics remains an open challenge. In this work, we investigate a fundamental question: \textit{Can a simple, unified 4D generative model learn the full spatiotemporal distribution directly?} To this end, we introduce \textbf{CardioDiT}, the first fully 4D diffusion transformer for CMR. CardioDiT first encodes 2D+t slices into a compact latent representation using a vector-quantized VAE, and then models the full 3D+t volume in latent space with a diffusion transformer \cite{dit} operating on the complete 4D tensor. By not factorizing space and time, the model learns globally consistent anatomy and temporally coherent cardiac motion without auxiliary constraints. We compare CardioDiT to baselines with progressively stronger spatiotemporal coupling: (i) a slice-wise 2D+t LDM conditioned on slice index, and (ii) a channel-merged 3D+t U-Net \cite{3d_t_ldm}. This controlled study highlights the importance of full 4D coupling. Our contributions are threefold: (1) formulation of CMR generation as direct modeling of the full 3D+t distribution, (2) the CardioDiT framework operating on complete 4D latent volumes, and (3) systematic analysis demonstrating improved anatomical and temporal consistency with fully 4D modeling.
\section{Methodology}
\begin{figure}[t]
    \centering
    \includegraphics[width=0.9\textwidth]{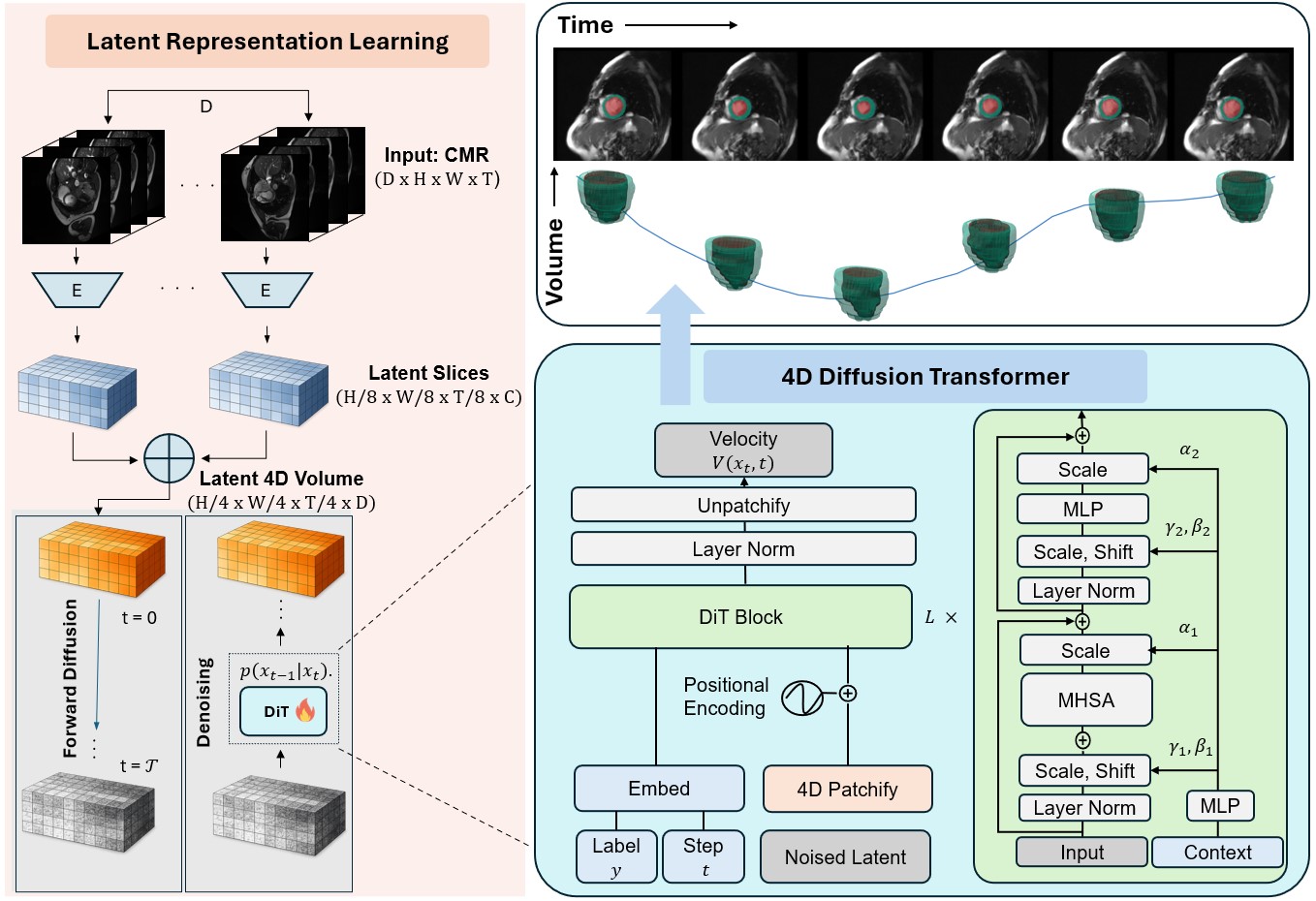}
    \caption{CardioDiT framework for 4D CMR synthesis. A VQ-GAN encodes 2D+t slices into latents stacked along $d$ to form a 4D representation. A 4D Diffusion Transformer denoises the noisy latent, followed by VQ-GAN decoding to reconstruct the full volume.}
    \label{fig:abstract}
\end{figure}

This work builds on our concurrently proposed 3D Diffusion Transformer architecture (see Supp. 1) and extends it to handle spatiotemporal data in the 3D+t setting for CMR. Specifically, we introduce 4D positional encodings and a 4D patchification strategy, enabling the model to explicitly capture temporal dynamics in addition to volumetric spatial features. Unlike the 3D approach, which focuses solely on static 3D volumes, the proposed 4D architecture is capable of modeling cardiac motion over time, allowing both unconditional and potentially conditional synthesis of temporally coherent CMR images (Fig. \ref{fig:abstract}).
\subsection{Spatiotemporal Latent Autoencoder}
We first train a spatio-temporal VQ-GAN to learn a compact latent representation of CMR volumes. Given a batch of CMR volumes  $( c, d, h, w, t)$ (channels, depth, height, width, temporal frames), we adopt a 3D VQ-GAN that randomly slices along the $d$-axis and reconstructs the $(h, w, t)$ dimensions. After training, we encode the full $3D+t$ volumes by sequentially slicing along $D$, encoding each slice into latent tokens, and stacking the latent representations to reconstruct the full $3D+t$ latent volume. 
\subsection{4D Diffusion Transformer }
Next, we train a diffusion transformer to model the latent 4D CMR representations. Latent volumes are patchified into 4D patches of size $(p_d, p_h, p_w, p_t) = (1,4,4,2)$ and linearly projected into token embeddings. These embeddings are combined with 4D sine-cosine positional encodings to preserve spatial and temporal structure: Given a latent volume $\mathbf{z} \in \mathbb{R}^{d \times h \times w \times t}$, the patchification and embedding process produces tokens $\mathbf{x} \in \mathbb{R}^{N \times e}$, where $N$ is the number of 4D patches and $e$ the embedding dimension. 
To enable efficient training with long token sequences, we employ FlashAttention, which reduces memory complexity to $\mathcal{O}(N)$ \cite{flashattention}. 
\subsection{Datasets}
\textbf{Public:} Cine short-axis CMR from \acrfull{acdc}~\cite{bernard2018acdc} and \acrfull{mm2}~\cite{campello2021mnm2,martinisla2023mnm2} were used. For \acrshort{acdc}, 100 cases were used for training and 50 for validation; for \acrshort{mm2}, 289 for training and 58 for validation. All volumes were resampled to the median spatial resolution, center-cropped, or padded to $6 \times 256 \times 256$ $(d,h,w)$. The temporal dimension was standardized to 32 frames via cyclic repetition.\\
\textbf{Private:}
A private CMR dataset was collected at Clinic \textit{anonym} using three different scanners. 813 samples were used for training and 102 for validation. All volumes were resampled to the median spatial resolution of corresponding scanners, cropped or padded to a fixed resolution of $10 \times 224 \times 224 \times 32$, matching the preprocessing applied to the public datasets.
\subsection{Experimental Setup}
\begin{figure}[!t]
    \centering
    \includegraphics[width=\textwidth]{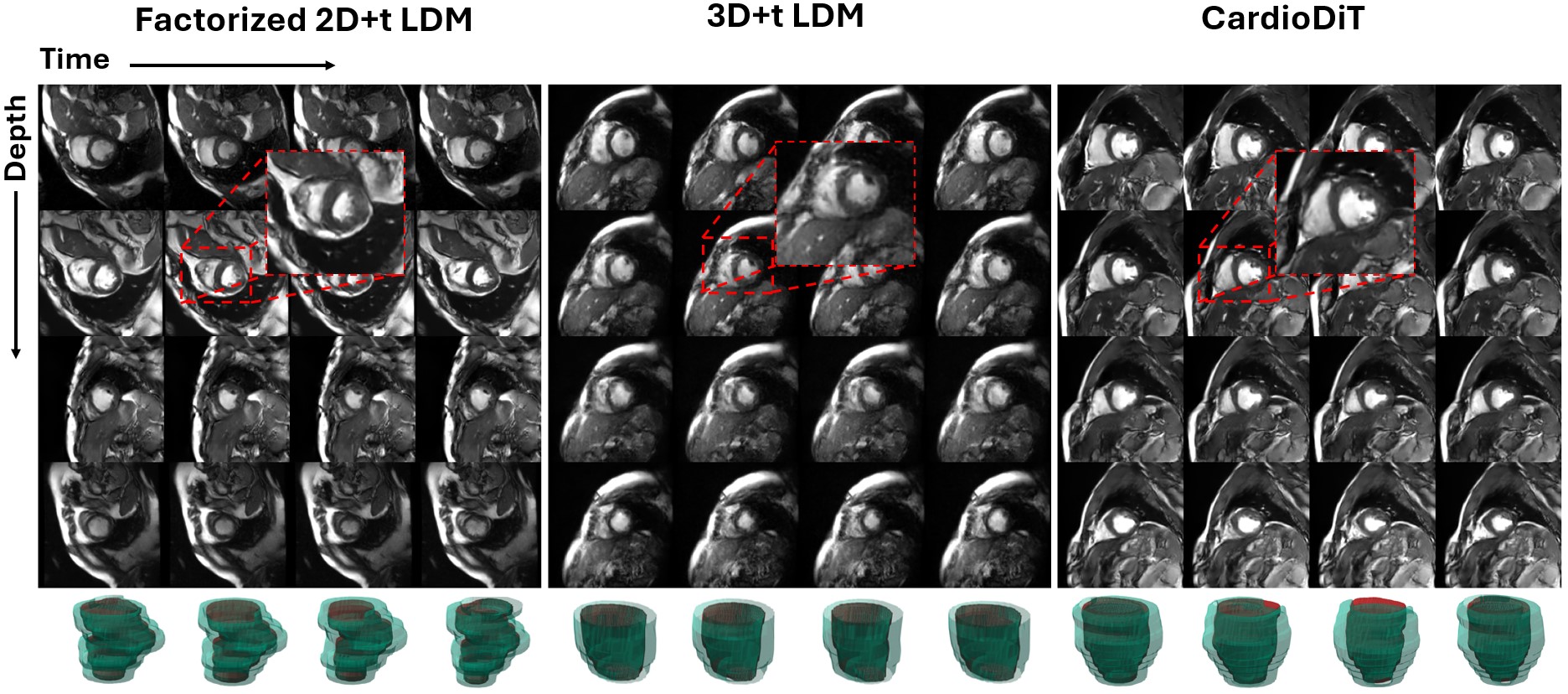}
    \caption{Visualized slices across time and depth for one synthetic sample per model, with corresponding LV meshes. Noteworthy are strong inter-slice discontinuities for the 2D+t LDM, blurry anatomical boundaries for the 3D+t U-Net, and anatomical consistency  for CardioDiT.}
    \label{fig:qualitative}
\end{figure}
To assess the importance of explicit 4D modeling, we compare CardioDiT against representative baselines with progressively stronger spatiotemporal coupling:\\
\textbf{Factorized 2D+t LDM:} A U-Net–based 2D+t latent diffusion model conditioned on slice index $s$, modeling each short-axis slice independently as $p(X)=\prod_{s=1}^{D} p(x_{s,1:T}\mid s)$, where $x_{s,1:T}$ denotes the temporal sequence at slice $s$. Spatial self-attention is applied at the two lowest-resolution stages.\\
\textbf{Channel-merged 3D+t LDM:} Following \cite{3d_t_ldm}, slice latents are concatenated along the channel dimension and processed jointly using a 3D U-Net, yielding $p(X)=p_{3Dconv}(x_{1:D,1:H,1:W,1:T})$. Spatial self-attention is likewise used at the two lowest-resolution stages. Both architectures follow implementations of \cite{medlord}. \\
\textbf{CardioDiT (ours):} CardioDiT models the complete 4D distribution $p(X) = p_{4D}(x_{1:D,1:H,1:W,1:T})$ jointly in latent space using a diffusion transformer.\\
We qualitatively and quantitatively evaluate the generated CMR volumes using complementary metrics for spatial consistency, distributional similarity, and temporal cardiac function.
To evaluate physiological plausibility, we analyze left ventricular (LV) volume curves across time. For each generated and real sample, we compute LV cavity volumes using a segmentation model trained with the nnU-Net framework~\cite{Isensee2024nnUNet} on the\acrshort{mm2}~\cite{campello2021mnm2,martinisla2023mnm2} training cohort.
To evaluate through-plane anatomical consistency, we compute the \textit{$d$-SSIM}, defined as the structural similarity between consecutive slices along the short-axis ($d$) direction for each time point \cite{ssim}. For each volume, SSIM values are aggregated across both spatial ($d$) and temporal ($t$) dimensions, and mean across all samples is reported. In addition, we quantify geometric consistency of the left ventricle at end-diastole (ED). For each slice, the center of mass of the LV segmentation is computed, and a linear regression line is fitted through these points along the through-plane direction. The mean absolute residual $AR_{ED}$ from this fitted axis is then reported as a measure of inter-slice alignment. Analogous to $d$-SSIM, performance is assessed relative to the real reference distribution, where values closer to the reference indicate better anatomical alignment.
For distributional similarity, we compute the Fréchet Inception Distance (FID) using the pretrained foundation model \textit{CineMA}, trained on short-axis CMR \cite{cinema}. We extract features for each CMR volume across all time points and compute FID, precision and recall \cite{precisionrecall}. The neighborhood parameter $k$ for precision and recall is selected such that a real--real comparison yields scores greater than 0.95, ensuring a stable and meaningful operating point. All comparisons are performed against the complete set of real validation samples for $N=50$ synthetic samples.
To assess temporal motion modeling, we report the mean and standard deviation of the ejection fraction (EF), defined as
$\text{EF} = \frac{V_{\text{ED}} - V_{\text{ES}}}{V_{\text{ED}}}$,
where $V_{\text{ED}}$ and $V_{\text{ES}}$ denote the end-diastolic and end-systolic volumes, respectively,
and quantify distributional differences in cardiac function by computing the Wasserstein distance ($W_2$) between the EF distributions of synthetic and real validation samples \cite{wasserstein2}. Additionally, we analyze the scalability of CardioDiT by evaluating three CardioDiT variants with increasing numbers of DiT blocks $(S,B,L) = (8,12,16)$ on the private dataset. 
The autoencoder was trained for 500 epochs with a compression rate factor $f = 8$ in all dimensions, codebook size $c' = 4096$ and embedding dimension $emb = 8$. CardioDiT was trained for 600,000 iterations using a cosine noise schedule with $T_{\text{diff}} = 300$ timesteps, AdamW optimizer and a fixed learning rate of $LR = 10e-4$. CardioDiT can run on 24GB VRAM.
\begin{table*}[!b]
\centering
\caption{Quantitative comparison of CMR generation models using distributional (FID, Precision, Recall), structural ($d$-SSIM, AR$_{\text{ED}}$) and clinical EF distribution statistics. Lower is better $\downarrow$, higher is better $\uparrow$, closest to real is better $\sim$.}
\label{tab:quantitative_results_merged}
\begin{tabular}{lccccccc}
\toprule
Model & FID $\downarrow$ & Precision $\uparrow$ & Recall $\uparrow$ & $d$-SSIM $\sim$ & AR$_{\text{ED}}$ $\sim$  & $\overline{EF}$ (Std) $\sim$  & $W_2$ $\downarrow$ \\
\midrule
\multicolumn{8}{c}{\textit{Public Dataset}} \\
Real & 52.8 & 0.95 & 0.96 & 0.55 (0.10) & 2.1 (1.0) & 47.6 (17.1) & 3.09  \\
Fact. 2D+t LDM & 100.1 & 0.89 & 0.17 & 0.39 (0.04) & 5.1 (1.1) & 45.3 (8.0)$^\star$ & 8.90  \\
3D+t U-Net & 221.5 & 0.59 & 0.64 & 0.75 (0.10) & 1.0 (0.3) & 44.5 (10.0) & 7.14 \\
\textbf{CardioDiT} & \textbf{71.9} & \textbf{0.97} & \textbf{0.70} & \textbf{0.69 (0.10)} & \textbf{1.3 (0.5)} & \textbf{49.3 (15.1)} & \textbf{2.67} \\
\midrule
\multicolumn{8}{c}{\textit{Private Dataset}} \\
Real & 11.4 & 0.92 & 0.92 & 0.63 (0.12) & 1.9 (0.7) & 44.6 (7.3) & 2.11 \\
Fact. 2D+t LDM & 32.9 & 0.58 & 0.36 & 0.50 (0.04) &5.0 (1.0)  & 41.1 (6.5)$^\star$ & 3.51 \\
3D+t U-Net & 51.3 & 0.43 & 0.23 & 0.73 (0.03) & 1.1(0.3) & 38.1 (5.1) & 7.23 \\
\textbf{CardioDiT} & \textbf{21.2} & \textbf{0.82} & \textbf{0.54} & \textbf{0.69 (0.09)} & \textbf{1.7 (0.4)} & \textbf{42.3 (6.3)} & \textbf{3.05} \\
\bottomrule
\end{tabular}
{\footnotesize 
$^\star$ EF is a clinically meaningful volumetric measurement. Inter-slice discontinuities in the 2D+t LDM make the computed EF physiologically uninterpretable.}
\end{table*}
\section{Results}
\subsection{Visual Assessment of 4D Consistency}
We qualitatively assess spatial fidelity and temporal coherence across models. Fig.~\ref{fig:qualitative} shows a representative CMR volume per model spanning the cardiac cycle. 
The factorized 2D+t LDM captures temporally plausible contraction and relaxation patterns within individual slices. Across depth, however, there are strong inter-slice discontinuities, leading to inconsistent ventricular structures despite roughly correct global positioning of the left ventricle. 
The 3D+t U-Net improves volumetric consistency, producing smoother structures across slices while maintaining reasonable temporal dynamics. Nevertheless, image fidelity is reduced, with blurred anatomical boundaries and diminished high-frequency detail.
In contrast, CardioDiT generates anatomically consistent ventricular structures across slices, maintains smooth and physiologically plausible motion across time, and preserves fine anatomical detail.
\subsection{Quantitative Evaluation}
As shown in Tab. \ref{tab:quantitative_results_merged} CardioDiT achieves the best FID and precision on both datasets, while substantially improving recall compared to the baselines. This indicates that our model generates high-fidelity samples while capturing a broader portion of the real data distribution. CardioDiT achieves $d$-SSIM  and $AR_{ED}$ closest to the real images among the evaluated models, indicating superior anatomical consistency across the through-plane direction, which is consistent with the qualitative results in Fig.~\ref{fig:qualitative}. For clinical EF statistics, CardioDiT achieves the closest match to the real EF distribution, reflected in both mean/standard deviation and the lowest $W_2$.
While the factorized 2D+t LDM yields EF statistics within a comparable range, the computation of EF does not explicitly enforce anatomical consistency in the through-plane direction. EF is a clinically meaningful volumetric measurement, and inter-slice discontinuities in the 2D+t LDM make the computed values physiologically uninterpretable. 
\begin{figure}[!t]
    \centering
    \includegraphics[width=0.85\textwidth]{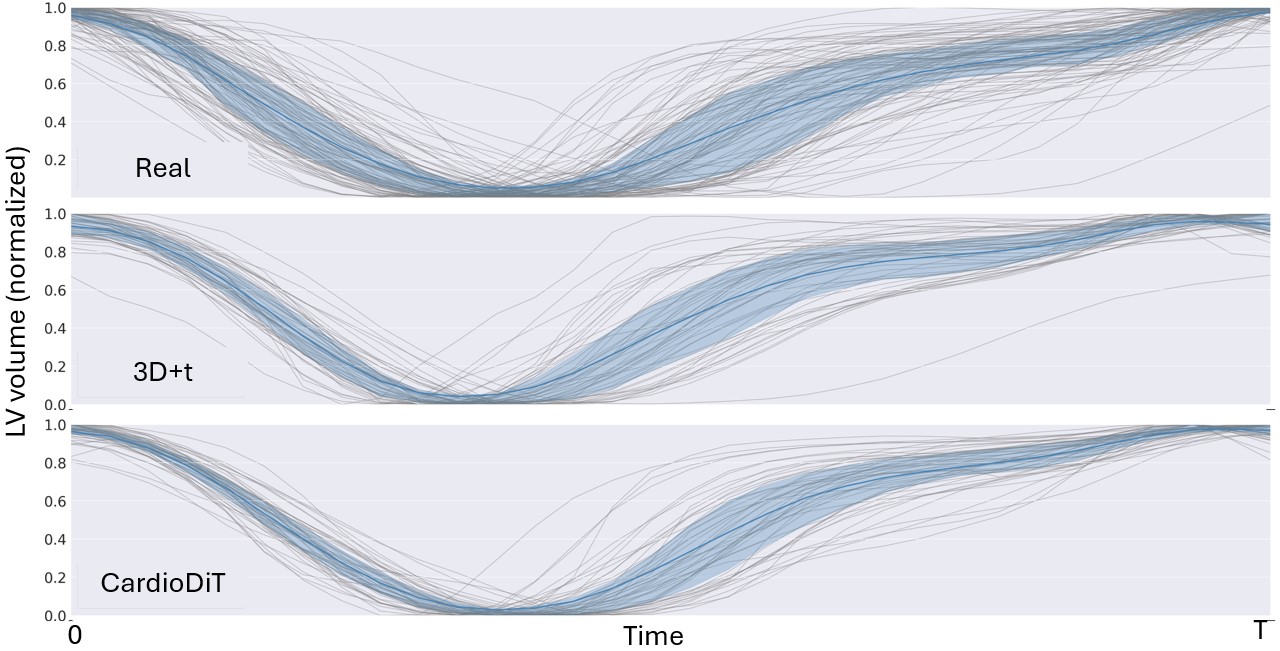}
    \caption{Normalized left ventricular volume-time curves derived from cine CMR blood pool segmentation, temporally aligned to peak-to-peak volume and scaled to equal length $T$. The mean curve (dark blue), inter-quartile range (IQR, light blue), and individual subject curves (light gray) are shown for the studied generative models.}
\label{fig:volume_curves}
\end{figure}
Additionally, Fig.~\ref{fig:volume_curves} shows the mean normalized LV volume curves across the cardiac cycle for the volumetric models. Both CardioDiT and the 3D+t LDM capture the overall temporal trend, with mean curves and interquartile ranges closely following the real data. Differences become apparent at the individual curve level, particularly during the transition from end-systole to early/mid-diastole, where CardioDiT exhibits variability more closely aligned with the real distribution. Volume curves for the factorized 2D+t LDM are not reported, as it requires anatomically consistent 3D structure to be clinically interpretable. Models that do not enforce through-plane coherence cannot yield physiologically meaningful volumetric curves. Overall, while both volumetric models reproduce cardiac motion dynamics, CardioDiT achieves the most consistent performance across fidelity, distributional alignment, and temporal variability.
\paragraph{Architecture Ablation Study}
As shown in Tab. \ref{tab:ablation_results} increasing model capacity consistently improves generative quality, as reflected by the monotonic decrease in FID from CardioDiT-S to CardioDiT-L. Slice consistency ($d$-SSIM) remains stable across scales. Notably, CardioDiT-S, despite having substantially fewer parameters, achieves better scores than all baseline methods in Tab.~\ref{tab:quantitative_results_merged}, demonstrating the efficiency of the proposed architecture.
Clinical EF statistics remain comparable across model sizes,  suggesting that scaling primarily enhances distributional fidelity rather than altering cardiac motion properties. 

\section{Discussion}
In this work, we investigated whether fully joint 4D transformer-based modeling can be used for cine CMR synthesis. By directly learning the full 3D+t distribution with a diffusion transformer backbone, CardioDiT achieves coherent cardiac anatomy and physiologically plausible motion dynamics using a conceptually simple, globally coupled architecture. Quantitative and qualitative results demonstrate improved through-plane consistency compared to slice-wise and channel-merged baselines, while temporal analysis confirms smooth contraction–relaxation dynamics and realistic ejection fraction distributions. These findings indicate that jointly modeling space and time encourages physiologically meaningful motion patterns to emerge implicitly.
A key design principle of our approach is architectural simplicity: rather than introducing modality-specific temporal mechanisms, we rely on a unified transformer operating on 4D latent tokens. The ablation study shows that increasing model capacity consistently improves spatial fidelity, suggesting that once spatiotemporal tokens are learned jointly, performance gains primarily arise from scaling rather than additional structural constraints \cite{dit}.
Although this study focused on unconditional synthesis, the framework naturally supports conditional generation. Conditioning signals can be incorporated via adaptive layer normalization or lightweight token adapters, enabling integration of demographic, pathological, or spatial priors without modifying the core architecture (see Supp. 1).
More broadly, operating in a unified spatiotemporal latent space aligns CardioDiT with emerging foundation-model principles, facilitating integration with multimodal systems or downstream analysis pipelines \cite{meddit_agent}. Beyond cine CMR, the proposed fully joint 4D modeling framework is applicable to other 4D imaging domains, including settings with a substantially higher number of d-slices. 
Limitations include the use of temporal standardization via cyclic repetition, which may introduce artificial periodicity. Future work should explore variable-length formulations to better capture physiological timing variability.
\begin{table}[!t]
\centering
\caption{Scalability of CardioDiT. Inference time $T$, VRAM and number of parameters were measured for a single sample on a NVIDIA RTX 4090 GPU.}
\label{tab:ablation_results}
\begin{tabular}{lccccccc}
\toprule
Model & FID $\downarrow$ & $d$-SSIM $\sim$  & $\overline{EF}$ (Std) $\sim$ & $W_2$ $\downarrow$ & \#Par (M) &T(s) $\downarrow$  & VRAM(GB)$\downarrow$ \\
\midrule
CardioDiT-S  &30.5  & 0.68(0.06) & 41.8 (7.7) & 3.56 & 92.23  & 30 & \textbf{1.2}\\
CardioDiT-B  &25.9  & 0.69(0.06) & 41.4 (4.7) & 4.13 & 134.74 & 45 & 1.3\\
\textbf{CardioDiT-L} &\textbf{21.2}  & \textbf{0.69(0.09)} &  \textbf{42.3 (6.3)} & \textbf{3.05}& 177.24 & 60 & 1.5\\
 Fact. 2D+t  & 32.9 & 0.50(0.04) & 41.1 (6.5) & 3.51& 166 & 20 & 1.5\\
 3D+t  & 51.3 & 0.73(0.03)  & 38.1 (5.1) & 7.23 & 578 & \textbf{15} & 3.8\\
\bottomrule
\end{tabular}
\end{table}

%
%

\subsubsection{\ackname}
This work was supported by Heidelberg Faculty of Medicine at Heidelberg University, by the Multi-DimensionAI project of the Carl Zeiss Foundation (P2022-08-010) and by the EU-Horizon Project \emph{DVPS} (101213369). The authors acknowledge \emph{1-} the data storage service SDS@hd supported by the Ministry of Science, Research and the Arts Baden-Württemberg (MWK) and the German Research Foundation (DFG) through grant INST 35/1314-1 FUGG and INST 35/1503-1 FUGG, \emph{2-} the state of Baden-Württemberg through bwHPC and the DFG through grant INST 35/1597-1 FUGG.

\bibliographystyle{splncs04}
\bibliography{4D_DiT.bib}

\end{document}